\pdfoutput=1
\documentclass[11pt]{article}
\usepackage{arabtex}
\setcode{utf8}
\usepackage{acl}

\usepackage{times}
\usepackage{latexsym}
\usepackage{tabularx}
\usepackage{algorithm}
\usepackage{algpseudocode}
\usepackage{array}
\usepackage{booktabs}
\usepackage{amsmath,amsfonts,amssymb}
\usepackage{float}
\usepackage{url}
\usepackage[T1]{fontenc}
\usepackage[utf8]{inputenc}

\usepackage{microtype}

\usepackage{inconsolata}

\usepackage{graphicx}
\title{Sharif-STR at SemEval-2024 Task 1: Transformer as a Regression Model for Fine-Grained Scoring of Textual Semantic Relations}

\author{
    \textbf{Seyedeh Fatemeh Ebrahimi}$^{\heartsuit}$, 
    \textbf{Karim Akhavan Azari}$^{\heartsuit}$, 
    \textbf{Amirmasoud Iravani}$^{\sharp}$ \\
    \textbf{Hadi Alizadeh}$^{\diamondsuit}$, 
    \textbf{Zeinab Sadat Taghavi}$^{\heartsuit}$, 
    \textbf{Hossein Sameti}$^{\heartsuit}$ \\
    Ferdowsi University of Mashhad, Mashhad, Iran$^{\sharp}$ \\
    Sharif University of Technology, Tehran, Iran$^{\heartsuit}$ \\
    Iran Broadcasting University, Tehran, Iran$^{\diamondsuit}$ \\
    \\
    \texttt{\{sfati.ebrahimi, karim.akhavan, zeinabtaghavi, sameti\}@sharif.edu} \\
    \texttt{a.iravani@mail.um.ac.ir} \\
    \texttt{alizadeh.hadi08@gmail.com} \\
}

\begin{document}
\maketitle
\begin{abstract}
Semantic Textual Relatedness holds significant relevance in Natural Language Processing, finding applications across various domains. Traditionally, approaches to STR have relied on knowledge-based and statistical methods. However, with the emergence of Large Language Models, there has been a paradigm shift, ushering in new methodologies. In this paper, we delve into the investigation of sentence-level STR within Track A (Supervised) by leveraging fine-tuning techniques on the RoBERTa transformer. Our study focuses on assessing the efficacy of this approach across different languages. Notably, our findings indicate promising advancements in STR performance, particularly in Latin languages. Specifically, our results demonstrate notable improvements in English, achieving a correlation of 0.82 and securing a commendable 19th rank. Similarly, in Spanish, we achieved a correlation of 0.67, securing the 15th position. However, our approach encounters challenges in languages like Arabic, where we observed a correlation of only 0.38, resulting in a 20th rank.
\end{abstract}

\section{Introduction}

STR delineates the meaningful association between linguistic units, showcasing conceptual proximity within a shared semantic frame \citep{HadjTaieb2019ASO, Abdalla2021WhatMS}. For instance, "cup" and "coffee" are related in meaning, yet they are not synonymous \citep{UBMA_285413791}. Despite its crucial role in various NLP applications such as Spelling Correction, Word Sense Disambiguation, Plagiarism Detection, Opinion Mining, and Information Retrieval \citep{FrancoSalvador2016ASS, Chen2017ASS, HadjTaieb2019ASO}, STR has garnered less attention compared to Semantic Textual Similarity (STS) due to a scarcity of available datasets.
Addressing this gap, \citet{Abdalla2021WhatMS}, and \citet{ousidhoum2024semrel2024} contributed to the field by constructing the first sentence-level STR datasets. In this paper, we endeavor to tackle the STR problem within shared Task 1\citep{ousidhoum-etal-2024-semeval}, Track A, leveraging supervised data in English, Spanish, and Arabic languages provided by \citet{ousidhoum2024semrel2024}. Additionally, we briefly explore Track C and provide supplementary details in Appendix B as a secondary objective.

Building upon the findings of \citet{Abdalla2021WhatMS}, which underscore the superior performance of fine-tuning Transformer models in supervised tasks, our proposed system captures the relationship among sentences by fine-tuning the RoBERTa Transformer \citep{Liu2019RoBERTaAR}. At the core of our system, we employ a pre-trained RoBERTa model as a regression model and fine-tune it to generate a floating-point value for the input text. During the pre-training process of RoBERTa, the emphasis is placed on tasks related to NLU. This involves exposing the model to a diverse range of linguistic contexts and training it to comprehend the nuances of language. Furthermore, the integration of a Classifier Head enables sentence classification, a pivotal aspect of our system architecture elaborated upon in section 3.

Our experimental results showcase promising performance on English and Spanish datasets, achieving respective correlation rates of 0.82 and 0.67 on test data, surpassing the baseline correlation set by SemEval-2024 at Subtask A \citep{ousidhoum-etal-2024-semeval}. However, the model's performance on Arabic data falls short, yielding only a 38\% correlation on development data. We attribute this discrepancy to differences in the underlying RoBERTa model and its training methodology across Latin and non-Latin languages, a topic further explored in section 5. To promote reproducibility and facilitate future research endeavors, the complete codebase of our project has been shared on GitHub\footnote{\url{https://github.com/Sharif-SLPL/Sharif-STR}}.

\section{Background}

\subsection{Dataset Overview}

The SemEval-2024 Task 1 is structured into Tracks A, B, and C, each tailored to specific methodologies and objectives. Our focus lies on Track A (Supervised), which utilizes labeled data to train STR systems. The datasets for Task 1 encompass training, development, and test sets across 14 languages, each comprising sentence pairs \citep{ousidhoum2024semrel2024}. Each sentence pair is annotated with a semantic relatedness score, ranging from 0 (indicating no relatedness) to 1 (suggesting strong relatedness). Participants are tasked with predicting the degree of semantic relatedness between sentence pairs, crucial for furthering research in NLP.

\subsection{Related Work}

The exploration of sentence-level STR has been hindered by the scarcity of available datasets \citep{Abdalla2021WhatMS}. Existing datasets, such as those compiled by \citet{Finkelstein2002PlacingSI}, \citet{Gurevych2006ThinkingBT}, \citet{Panchenko2016HumanAM}, and \citet{Asaadi2019BigBA}, predominantly focus on unigram and bigram STR. However, the seminal works of \citet{Abdalla2021WhatMS}, and \citet{ousidhoum2024semrel2024} paved the way for further research by constructing the first sentence-level STR datasets.
Traditionally, both STR and STS have been approached using knowledge-based and statistical methods \citep{Sadr2020ExploringTE, Chandrasekaran2020EvolutionOS}.
Notable efforts include the application of knowledge bases such as thesauri, ontologies, and dictionaries for STR, as surveyed by \citet{Salloum2020ASO}. Statistical methods, on the other hand, leverage features extracted from corpora, with prominent examples including Latent Dirichlet Allocation (LDA) by \citet{Blei2009LatentDA} and Latent Semantic Analysis (LSA) by \citet{Landauer2008LatentSA} for topic modeling.\newline

In recent years, the application of deep learning methodologies has surpassed traditional approaches in STS tasks. Noteworthy advancements include the Tree-LSTM model proposed by \citet{Tai2015ImprovedSR}, which outperformed other neural network models in SemEval-2014. \citet{He2016PairwiseWI} introduced a hybrid architecture of Bi-LSTM and CNN, outperforming the Tree-LSTM model on the SICK dataset. \citet{Wang2016SentenceSL} achieved state-of-the-art results using the Word2Vec embeddings model in both the QASent and the WikiQA datasets, while \citet{Shao2017HCTIAS} leveraged GloVe embeddings to achieve the third rank in SemEval-2017.\\

Several studies have demonstrated that fine-tuning transformer-based models achieves state-of-the-art in comprehending the semantics of textual data. The transformer model, first introduced by \citet{Vaswani2017AttentionIA}, employs attention mechanisms to capture word semantics. Later on, \citet{Devlin2019BERTPO} utilized it to create BERT word embeddings. Subsequently, XLNet, proposed by \citet{Yang2019XLNetGA}, surpassed BERT in performance. Consequently, \citet{Lan2019ALBERTAL} introduced ALBERT, which outperforms previous models. Additional transformer-based variations of BERT models include TinyBERT \citep{jiao-etal-2020-tinybert}, RoBERTa \citep{Liu2019RoBERTaAR}, and DistilBERT \citep{Sanh2019DistilBERTAD}. Also, \citet{Raffel2019ExploringTL} presented five distinct versions of the T5 transformer model, each varying in parameter size. Their work demonstrated that the performance of these pretrained models improves with larger datasets and enhanced computational resources.\\ 

\citet{Laskar2020ContextualizedEB} addressed sentence similarity modeling within an answer selection task. Through experiments conducted, they showed that fine-tuning RoBERTa model achieves state-of-the-art performance across datasets. \citet{Yang2020MeasurementOS} showcased that the RoBERTa-based model achieved superior performance compared to the BERT and XLNET models in a clinical STS task, achieving a Pearson Correlation of 0.90. Similarly, \citet{Huang2021hubAS} conducted a comparison of TF-IDF combined with various models including ALBERT, BERT, and RoBERTa for word similarity detection in sentence pairs within Task 2 of SemEval-2021. Their experimental findings substantiated that RoBERTa yielded superior results by 0.846 on the test data. \citet{nasib2023references} addressed reference validation task by employing BERT, SBERT, and RoBERTa. His study illustrated the efficacy of fine-tuning a RoBERTa-based model for text classification tasks, achieving state-of-the-art performance across multiple benchmark datasets. He emphasized that optimizing the model's performance involves activities such as hyperparameter tuning, regularization, and data augmentation.\\

\citet{Abdalla2021WhatMS} conducted an extensive investigation into semantic sentence representation methods, revealing that supervised methods utilizing contextual embeddings, particularly those fine-tuning BERT or RoBERTa, outperform other techniques, reaching a correlation of 0.83. Building upon these findings, we adopt fine-tuning RoBERTa as the primary strategy in this paper. Subsequent sections will detail our system architecture.

\section{System Overview}

In this section, we present a comprehensive overview of our system's architecture, outlining the key algorithms and modeling decisions that underpin our model.

\subsection{Core Algorithms and System Architecture}

Our system harnesses the Transformer architecture for its ability to capture long-range dependencies. At its core, we harness the power of a pre-trained RoBERTa model \citep{Liu2019RoBERTaAR} for regression analysis, tailoring its parameters to accurately predict a floating-point value from the input text. While RoBERTa isn't explicitly trained for sentence relatedness scoring, its training encompasses an understanding of the relatedness of sentences within discourse, rendering it suitable for our task.

During the pre-training process of RoBERTa, the emphasis is placed on tasks related to NLU. This involves exposing the model to a diverse range of linguistic contexts and training it to comprehend the nuances of language. Our word embeddings utilize an embedding matrix with a dimensionality of 768. Position embeddings and token type embeddings further contribute to the model's comprehension of sequential and contextual information within the input data.

The RobertaEncoder comprises a stack of 12 identical RobertaLayers, each employing a multi-head self-attention mechanism. This mechanism enables the model to concurrently absorb different parts of the input sequence, showing promise in analyzing similarities between various inputs. Following the attention mechanism are intermediate sub-layers and output sub-layers. The intermediate sub-layer employs a fully connected feed-forward network with a GELU activation function, while the output sub-layer is responsible for proper transformation and normalization of features.

The classification head, positioned after the encoder, is tasked with generating the final output for sequence classification. It consists of a linear layer with 768 input features, followed by a dropout layer to prevent over-fitting. An additional linear layer featuring a solitary output neuron enables binary classification. By viewing the problem as a regression task, the classifier yields a linear output designed for a singular class, producing a probabilistic value indicative of the relatedness between input sentences.

\subsection{Resources} 

For training our model, we relied on the dataset provided for SemEval-2024 Task 1 \citep{ousidhoum2024semrel2024}. In addition to the primary dataset, we augmented our training dataset using the T5 model \citep{Raffel2019ExploringTL}. By leveraging T5's paraphrasing capabilities, we explored data augmentation techniques for Track A on the training sets of our dataset but failed to achieve consistent results across experiments. While some experiments showed an increase in model accuracy, in other cases, it did not alter the results. Data augmentation consistently worked well only on the English dataset. More details about data augmentation results and our secondary investigation on Track C are provided in Appendix A and B.

By incorporating both the SemEval-2024 Task 1 dataset \citep{ousidhoum2024semrel2024} and augmented training data generated by T5, our approach benefits from a comprehensive and diverse set of resources, enabling robust training and evaluation of our STR model across multiple languages and textual domains. 

\subsection{System Challenges}

Augmenting the dataset for training set using T5 paraphrases posed several challenges. Firstly, while the primary dataset was labeled through collaborative human judgment, the augmented data lacked this human validation. This absence of human labeling for the augmented data may potentially impact its quality. Moreover, the augmentation process introduced alterations to the diversity of the data, presenting a challenge to maintaining the original data variety.

The decision to employ data augmentation exclusively for testing purposes raises concerns regarding its potential impact on model quality. Addressing these challenges associated with data augmentation is crucial for improving the efficacy of our model. Exploring solutions to mitigate these issues can enhance our approach to tackling the task at hand.

\section{Experimental Setup}
\subsection{Dataset}
The dataset statistics utilized for each language are presented in Table 1: 

\begin{table*}[htbp]
\centering
\label{tab:data_stats}
\begin{tabular}{@{}lcccc@{}}
\toprule
\textbf{Language/Split} & \textbf{Dataset} & \textbf{Train} & \textbf{Testset} & \textbf{Devset} \\ \midrule
English & 5752 & 4400 & 1101 & 251 \\
Spanish & 1702 & 1249 & 313 & 140 \\
Arabic & 1360 & 1009 & 252 & 97 \\ \bottomrule
\end{tabular}
\caption{Dataset Statistics}
\end{table*}
As shown in Table 1, approximately 0.8 of the Task 1 dataset is allocated for system training, while the remainder is reserved for evaluation. The limited availability of training data necessitates cautious consideration during testing, as the model's performance may be influenced by the scarcity of training instances. Additionally, the entire development set is utilized for model selection.

\subsection{Pre-processing and Hyper-Parameter Tuning}
A crucial aspect of our pre-processing involves converting the labels (scores) of each data instance to float values, ensuring compatibility with the model's expected input format. Furthermore, the input texts undergo tokenization using the RoBERTa-tokenizer both during training and inference.

Hyperparameter tuning plays a pivotal role in optimizing model performance. Our tuning process encompasses exploring various hyper-parameters, including learning rates in the range of [0.00001, 0.00003], dropout rates ranging from [0.1, 0.3], batch sizes spanning [4, 32], and token sizes from [32, 128]. Through iterative experimentation, we determined that a learning rate of 0.00003, a dropout rate of 0.1, a token size of 128, a batch size of 16, and a weight decay of 0.01 yield optimal results across all languages.

The selection of an appropriate token size is not solely based on computational considerations; rather, it is informed by dataset analysis. Upon examination, it became evident that the majority of data instances are predominantly short, aligning with our token size choice. Additionally, truncation during tokenization supports the chosen token size, ensuring efficient model training without sacrificing data representativeness.

\subsubsection{Mean Squared Error (MSE)}

Mean Squared Error quantifies the average of the squared differences between predicted and actual values. It is calculated using the formula:
\begin{equation}
\label{eq:example}
\begin{aligned}
MSE = \frac{1}{N} \sum_{i=1}^{N} (y_i - \hat{y}_i)^2 
\end{aligned}
\end{equation}
Where \( N \) is the number of instances, \( y_i \) is the true label, and \( \hat{y}_i \) is the predicted value.Additionally, Mean Absolute Error computes the average absolute differences between predicted and actual values.Moreover, the R-squared score assesses the proportion of variance in the dependent variable explained by the independent variable.

These evaluation measures collectively shed light on our regression model's performance in predicting the degree of relatedness between text samples. Using these metrics together enables the monitoring of the model’s performance and, hence, facilitates decisions on hyper-parameters, model selection, etc. The evaluation method and hyper-parameter choices remain consistent across all models and languages. For the analysis of results presented in Section 5, the obtained scores were discretized and categorized into five distinct ranges to enhance visual understanding.

\section{Results}

\subsection{Findings}
A direct comparison with previous models and datasets similar to this task is challenging due to our specific focus on fine-tuning the RoBERTa model and utilizing the dataset provided by \citet{ousidhoum2024semrel2024}. Drawing from the insights of \citet{Raffel2019ExploringTL} working on the STS dataset, it is evident that the performance of transformer models improves with larger training corpora and enhanced computational resources. \citet{Raffel2019ExploringTL} demonstrated that the RoBERTa transformer-based model achieved a Pearson correlation of 0.922, surpassing ERNIE 2.0, DistilBERT, and TinyBERT on STS dataset benchmarks. Conversely, ALBERT, XLNet, and T5-11B outperformed RoBERTa on the same task, achieving a Pearson correlation of 0.925. Therefore, we recommend conducting a benchmark study of top-performing transformer models like RoBERTa, ALBERT, XLNet, and T5-11B in future research endeavors. Using the official metric of Spearman Correlation proposed in SemEval-2024 Task 1 \citep{ousidhoum-etal-2024-semeval}, our system achieves the following scores on different data splits and languages:

\begin{table}[htbp]
\centering
\begin{tabular}{@{}lcc@{}}
\toprule
\textbf{Language/Split} & \textbf{Devset} & \textbf{Testset(Competition)} \\ \midrule
English & 0.83 & 0.82 \\
Spanish & 0.71 & 0.67 \\
Arabic & 0.32 & 0.38 \\ \bottomrule
\end{tabular}
\caption{Correlation Metric Scores}
\label{tab:correlation}
\end{table}

As shown in Table 2, Firstly, comparing the performance between English, Spanish, and Arabic models, we observe varying degrees of success. The English model demonstrates the highest Spearman Correlation scores, both on the development and test sets, with scores of 0.83 and 0.82, respectively. This indicates that the English model performs relatively well in capturing the semantic relatedness between text pairs. Similarly, the Spanish model also achieves respectable scores, albeit slightly lower, with scores of 0.71 on the development set and 0.67 on the test set. However, the Arabic model lags significantly behind, exhibiting notably lower scores of 0.32 on the development set and 0.38 on the test set.

The disparity in performance between the Arabic model and the English and Spanish models could be attributed to several factors. One possible explanation is the availability and quality of training data. The Arabic dataset may suffer from a scarcity of labeled instances, resulting in a less robust model. Additionally, linguistic and structural differences between Arabic and Latin languages may pose challenges for the model in accurately capturing semantic relatedness. This discrepancy underscores the importance of adequately addressing language-specific characteristics and challenges in model development.

Furthermore, the analysis of the Arabic model's performance on the test set reveals a noteworthy observation. Despite achieving a relatively low Spearman Correlation score, the model appears to disproportionately classify most inputs as highly related. This discrepancy suggests a potential limitation in the model’s ability to discern varying degrees of relatedness accurately. It implies that while the model may perform adequately in certain aspects, such as overall correlation with human annotations, it may struggle with nuanced interpretations of relatedness levels in real-world scenarios. The output of the model is provided in Appendix D.

\begin{figure*}
  \centering
  \includegraphics[width=1\linewidth]{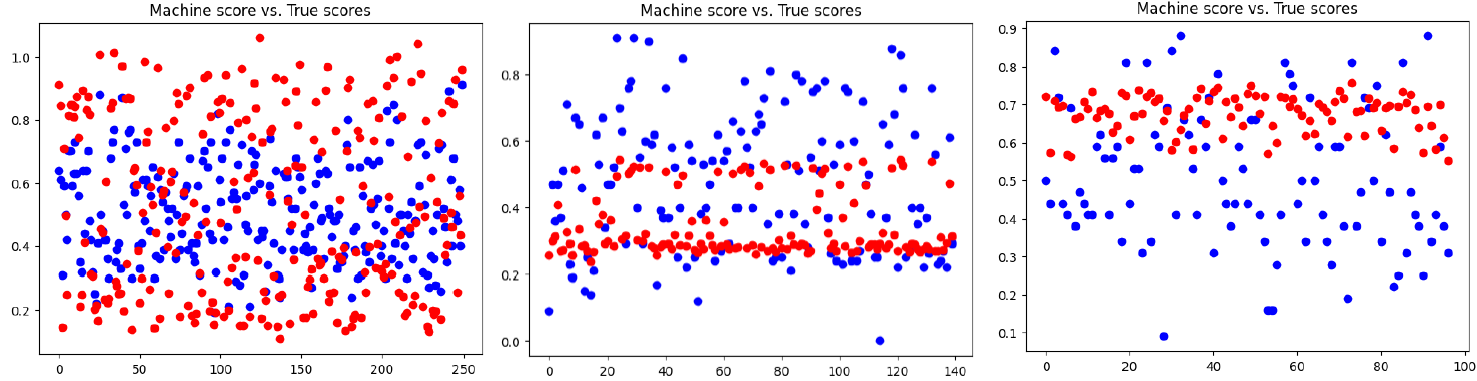}
  \caption{ Scatter Plots of English, Arabic and Spanish Languages }
\end{figure*}

\begin{figure*}
  \centering
  \includegraphics[width=1\linewidth]{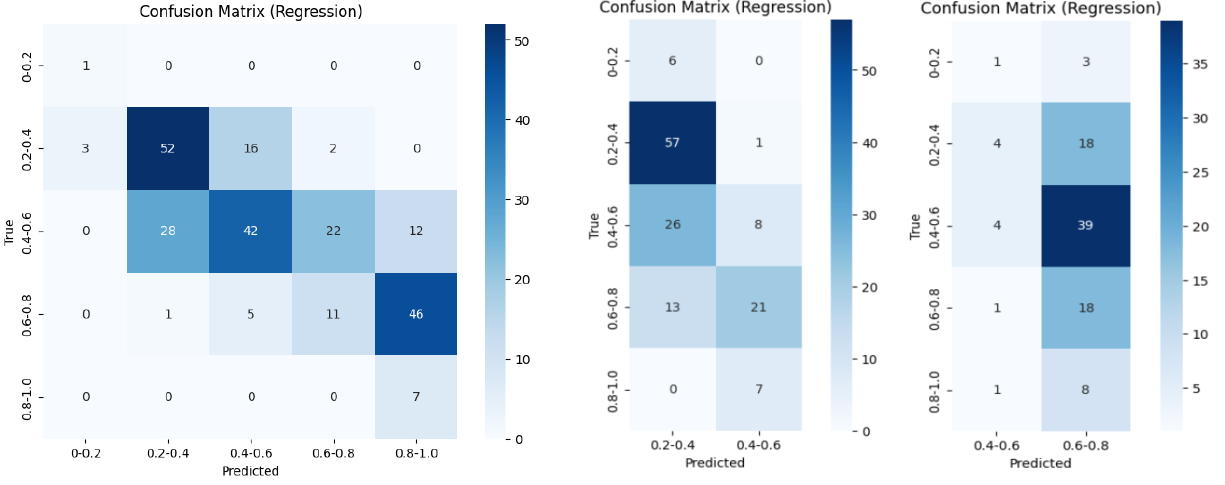}
  \caption{ The Confusion Matrix Plot of English, Arabic and Spanish Languages }
\end{figure*}

The scatter plots depicted in Figure 1, respectively for English, Spanish, and Arabic, illustrate the correlation between the model predictions and human annotations. The English model closely aligns with human annotations, while the Spanish model exhibits an even closer alignment on certain inputs. However, the Arabic model's performance varies, indicating discrepancies between predicted and actual relatedness scores. These findings underscore the importance of dataset size and linguistic nuances in model performance across different languages. Further investigation is warranted to elucidate the factors influencing model behavior and to improve performance, particularly in languages with limited training data.

\subsection{Error Analysis}

While confusion matrices are less commonly utilized in regression problems, discretizing the model's scores allows us to glean insights into its performance. Confusion matrix plots for English, Spanish, and Arabic are provided in Figure 2, respectively.
Upon examining the confusion matrix of the English dataset, it becomes apparent that the model performs well within certain score ranges. However, there are notable areas, particularly within the highly related range (0.6-1.0), where our model could benefit from improvement.

A similar observation holds true for the Spanish dataset, where the model demonstrates proficiency in predicting less related sentences but encounters challenges with highly related ones. Conversely, the Arabic dataset presents a markedly different scenario. While the majority of predictions fall within the mid-range of relatedness, they are predominantly incorrect. 

Based on the histogram and extracted statistics from the fine-tuning data in Figure 3 in Appendix C, it appears that the majority of the training data has a distribution centered around the median (Spanish Mean Score: 0.43, Arabic Mean Score: 0.50). Consequently, finte-tuned Arabic and Spanish models seem to have less capability in understanding data on both ends of the spectrum. 

These insights highlight the model's strengths and weaknesses across different datasets and underscore the need for further investigation into improving performance, particularly in accurately predicting highly related sentences across all languages. Further exploration of the factors contributing to model errors, such as dataset characteristics and linguistic nuances, is essential for refining the model's predictive capabilities.

\section{Conclusion}

In our investigation, we focused on fine-tuning RoBERTa for STR, primarily targeting Latin languages like English(0.82) and Spanish(0.67). While our approach showed promising results for these languages, particularly in achieving high correlation, the outlook was less favorable for Arabic(0.38). This echoes discussions in previous works, emphasizing the significant influence of the data on model performance. Our exploration into Track C, which is given in Appendix B, further enriched our understanding of the challenges and opportunities in STR system development. As a contribution to the field, we put forth several recommendations for enhancing STR systems. Firstly, we propose the development of additional Transformer models trained on diverse language families, focusing on languages that share similarities with Latin languages. Furthermore, a comprehensive benchmark of models on the STR dataset is essential, building on previous research that highlights the strong performance of models like ALBERT, XLNet, and T5-11B on the STS dataset. Moreover, the utilization of translation techniques and data augmentation methods could enhance model performance, particularly for languages with limited training data. In conclusion, our study sheds light on the nuances of STR system development and underscores the importance of considering language-specific factors and domain characteristics. By pursuing the avenues outlined in this paper, we aim to contribute to the advancement of STR research and facilitate the development of more robust and accurate models for NLU tasks.

\section*{Acknowledgments}
We express our gratitude to the Speech and Language Processing Laboratory at Sharif University of Technology\footnote{\url{https://github.com/Sharif-SLPL}} for offering us the opportunity for collaborative work.

\bibliography{custom}
\clearpage
\appendix
\section{Data Augmentation Results}
\label{Data augmentation} 
As we describe data augmentation in section 3.2, we use T5 model to augment some training data and use them in training of model.So in this section we show results of data augmentation effect on Pearson Correlation for English language in Table 3.

\begin{table}[H]
\resizebox{\columnwidth}{!}{%
\begin{tabular}{cc|c|c}
\hline
\multicolumn{2}{c|}{\begin{tabular}[c]{@{}c@{}}Model \\ hyper parameters\end{tabular}} & \begin{tabular}[c]{@{}c@{}}without\\ data augmentation\end{tabular} & \begin{tabular}[c]{@{}c@{}}with\\ data augmentation\end{tabular} \\ \hline
\multicolumn{1}{c|}{Learning rate}                        & 3e-5& 
                &                                                \\ \cline{1-2}
\multicolumn{1}{c|}{Max length}                           & 128& {0.79}                                                               & {0.81}                                                         \\ \cline{1-2}
\multicolumn{1}{c|}{Batch size}                           & 16 &                                                                     &                                                                  \\ \cline{1-2}
\multicolumn{1}{c|}{Epoch}                                & 4 &                                                                     &                                                                  \\ \hline
\end{tabular}%
}
\begin{center}
\caption{Data Augmentation Affect on Pearson Correlation}  
\end{center}

\end{table}
%~\ref{sec:appendix}
%\label{sec:appendix}

\section{ Track C - Cross-Lingual}
Using the translation method in Track C, we employed our Track A model trained on English language. The input sentences were first translated into English using the Google Translate API, followed by the utilization of the trained Track A model. The evaluation results demonstrate promising performance across some languages with this approach. However, errors might arise from either the Google Translate API or the model itself. Exploring alternative translation APIs could potentially enhance the overall performance. Figures 3, 4, and 5 display the outputs in Afrikaans, Amharic, and Modern Standard Arabic. Additionally, the high-quality output images are provided in our GitHub project.

\begin{table}[H]
\resizebox{\columnwidth}{!}{%
\begin{tabular}{l|cccc}
\hline
Test Data & Pearson Correlation & MSE \\
\hline
afr\_test\_with\_labels.csv & 0.8 & 0.0204 \\
amh\_test\_with\_labels.csv & 0.73 & 0.0309 \\
arb\_test\_with\_labels.csv & 0.51 & 0.0431 \\
\hline
\end{tabular}%
}
\caption{Track C Results}
\label{tab:track_c_results}
\end{table}

\begin{figure}[!htb]
  \centering
  \includegraphics[width=1\linewidth]{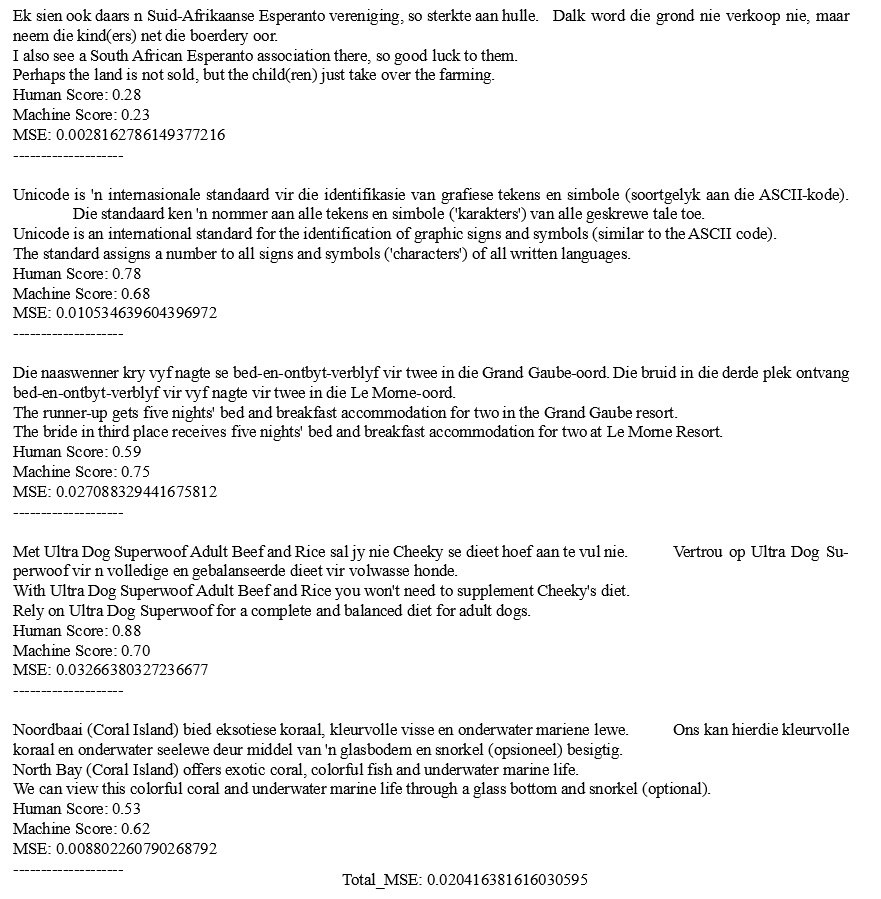}
  \caption{ Output of Afrikaans}
\end{figure}
\begin{figure}[!htb]
  \centering
  \includegraphics[width=1\linewidth]{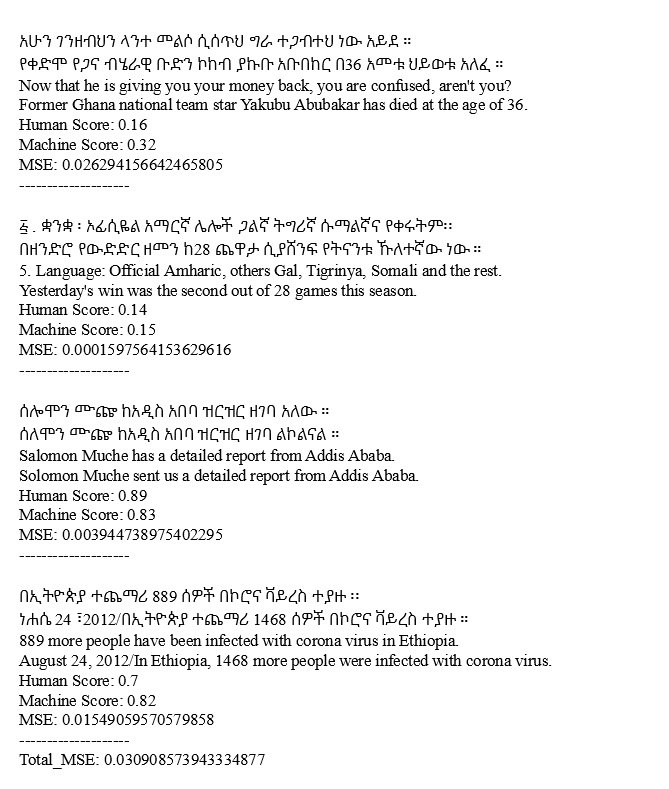}
  \caption{ Output of Amharic Language}
\end{figure}
\begin{figure}[!htb]
  \centering
  \includegraphics[width=1\linewidth]{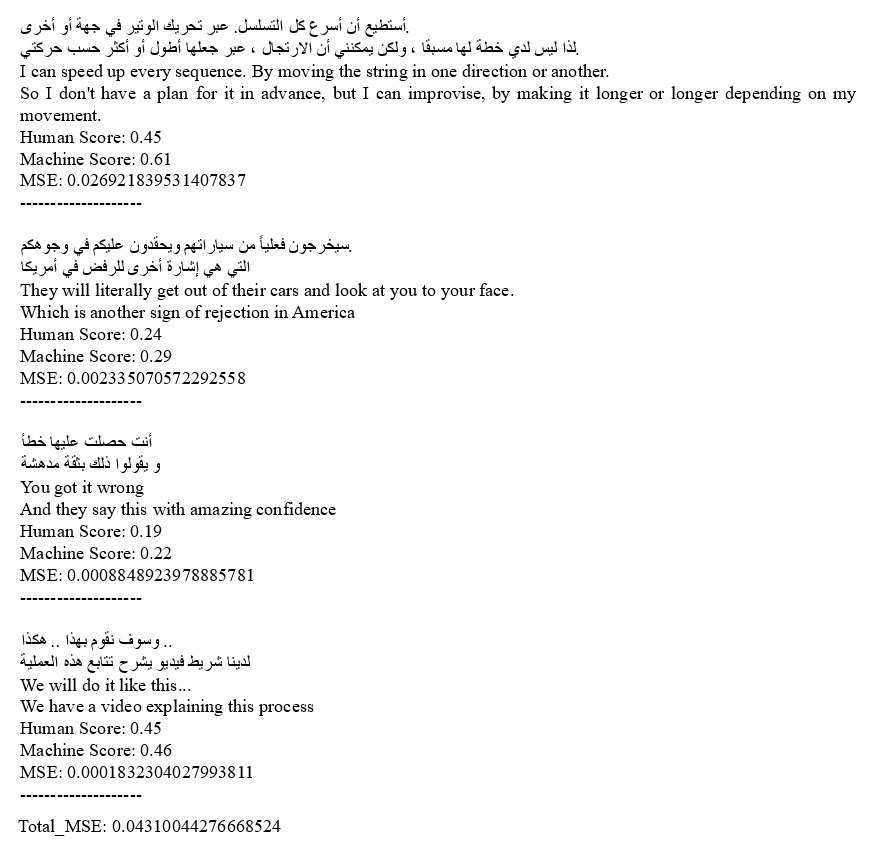}
  \caption{ Output of Modern Standard Arabic}
\end{figure}

\section{Histogram of Spanish and Arabic Languages}
\begin{figure}[H]
  \centering
  \includegraphics[width=1\linewidth]{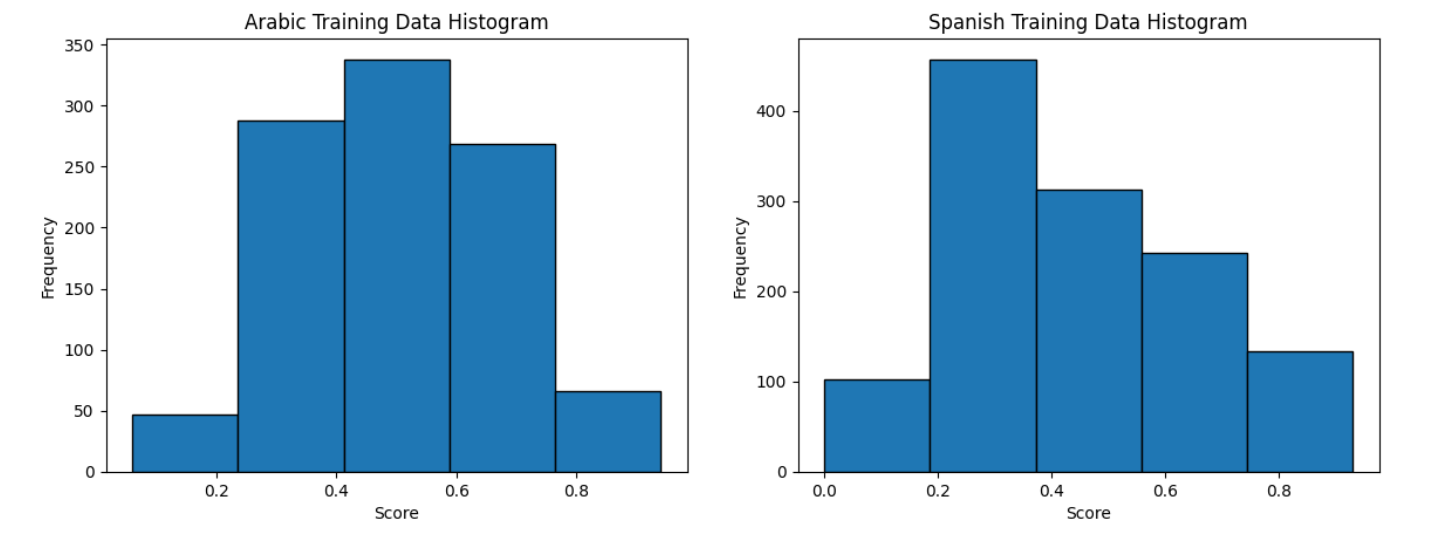}
  \caption{ Histogram of Spanish and Arabic Training Dataset}
\end{figure}

\section{ Outputs of Track A (Supervised)}
\begin{figure}[H]
  \centering
  \includegraphics[width=1\linewidth]{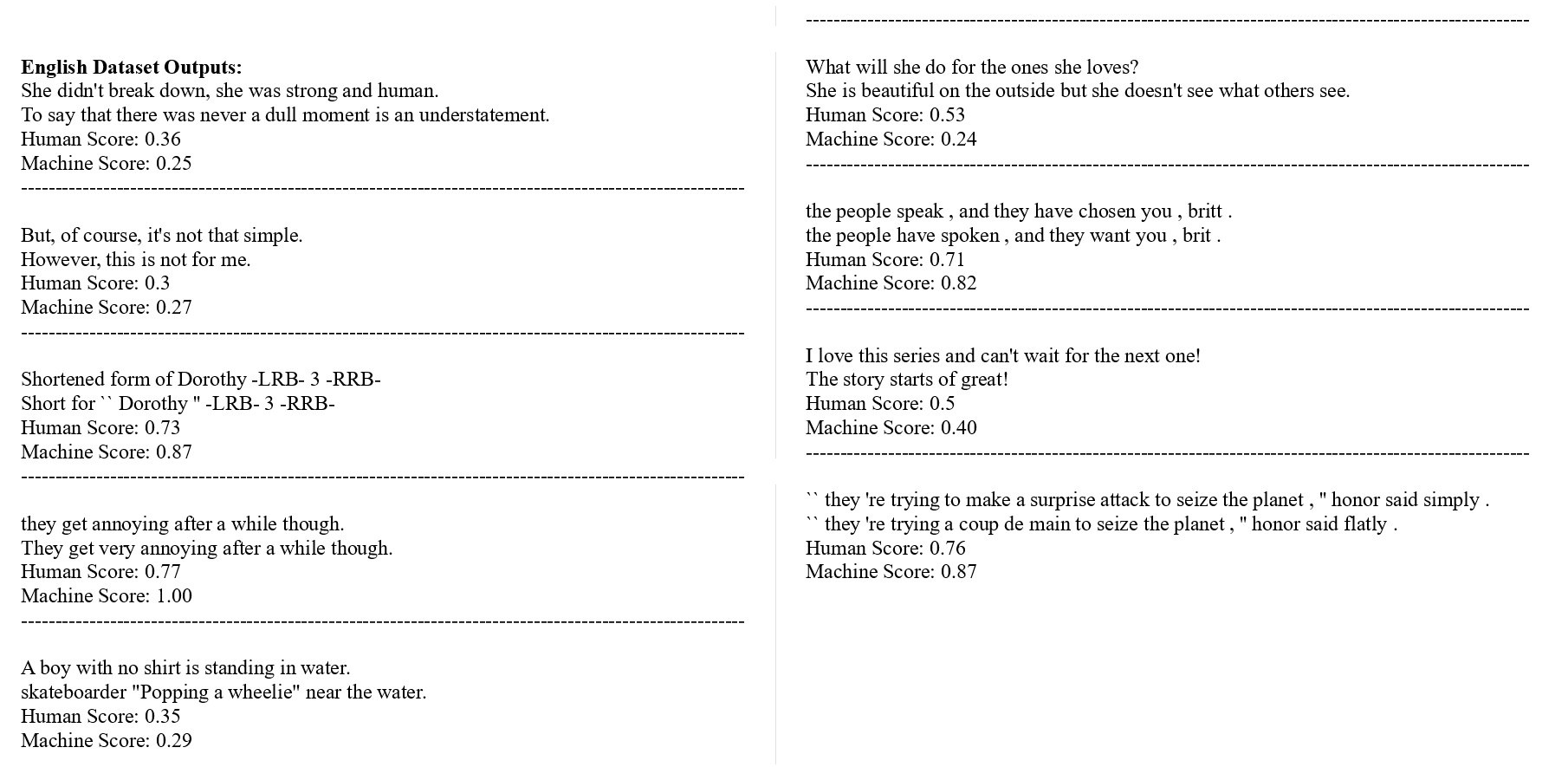}
  \caption{ Output of English Language }
\end{figure}

\begin{figure}[H]
  \centering
   \includegraphics[width=1\linewidth]{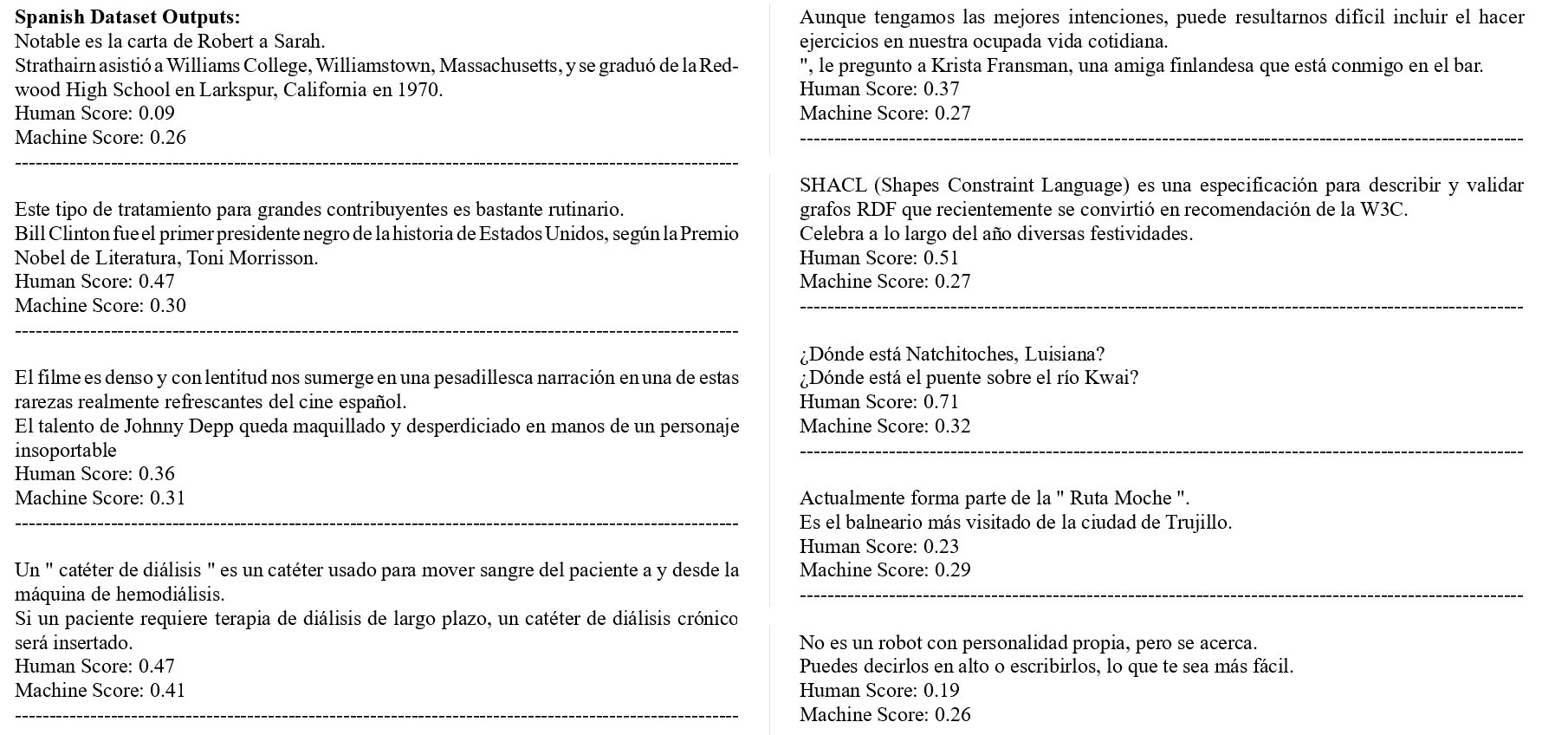}
   \caption{ Output of Spanish Language }
 \end{figure}
  \begin{figure}[H]
  \centering
   \includegraphics[width=1\linewidth]{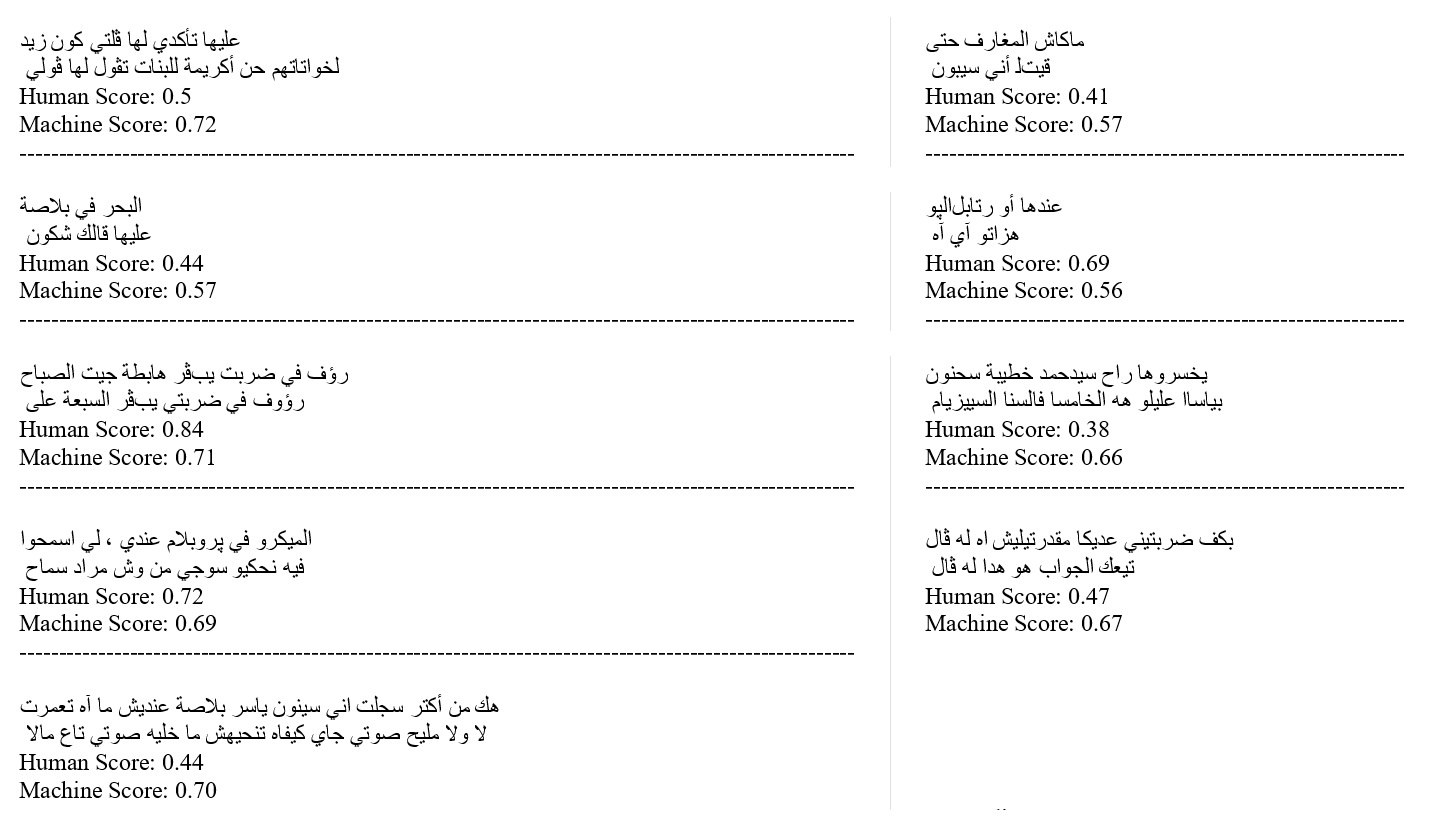}
   \caption{ Output of Arabic Language }
 \end{figure}

\end{document}